\documentclass[11pt,a4paper]{article}
\usepackage[hyperref]{acl2019}
\usepackage{times}
\usepackage{latexsym}

\usepackage{url}
\usepackage{wrapfig}
\usepackage{graphicx}
\usepackage{enumitem}
\usepackage{amsfonts}
\usepackage{amsmath}
\usepackage{multirow}
\usepackage[justification=centering, compatibility=false]{caption}
\usepackage{subcaption}
\usepackage{booktabs}
\usepackage[T1]{fontenc}

\aclfinalcopy 
\def\aclpaperid{1618} 

\title{Psycholinguistics meets Continual Learning: \\Measuring Catastrophic Forgetting in Visual Question Answering} 

\author{
Claudio Greco$^{1}$\\
\texttt{claudio.greco@unitn.it}
\And
Barbara Plank$^{2}$\\
\texttt{bplank@itu.dk}
\AND
Raquel Fern\'{a}ndez$^{3}$\\
\texttt{raquel.fernandez@uva.nl}
\And
Raffaella Bernardi$^{1,4}$\\
\texttt{raffaella.bernardi@unitn.it}
\AND
$^{1}$\normalfont CIMeC and $^{4}$\normalfont DISI \\University of Trento\And
$^{2}$\normalfont Dept. of Computer Science\\IT University of Copenhagen\And
$^{3}$\normalfont ILLC\\University of Amsterdam
} 

\date{}

\newcounter{tbsnr}
\newenvironment{tbs}
{\addtocounter{tbsnr}{1}\par\bigskip\noindent\fbox{\thetbsnr}
\hspace*{\fill}\begin{minipage}{7cm}\tt}
{\end{minipage}\hspace*{\fill}\bigskip}

\newcommand{\cut}[1]{}

\begin{document}
\maketitle

\begin{abstract}
We study the issue of catastrophic forgetting in the context of neural multimodal approaches to Visual Question Answering (VQA). Motivated by evidence from psycholinguistics, we devise a set of linguistically-informed VQA tasks, which differ by the types of questions involved (Wh-questions and polar questions). We test what impact task difficulty has on continual learning, and whether the order in which a child acquires question types facilitates computational models. Our results show that dramatic forgetting is at play and that task difficulty and order matter. Two well-known current continual learning methods mitigate the problem only to a limiting degree.
\end{abstract}


\section{Introduction}
\label{sec:introduction}
Supervised machine learning models are incapable of continuously learning new tasks,
as they forget how to perform the previously learned ones. This problem,
called \textit{catastrophic forgetting}, is prominent in artificial
neural networks \cite{mcclelland1995:there}. \textit{Continual
  Learning} (CL) addresses this problem by trying to equip models with
the capability to continuously learn new tasks over time~\cite{ring1997:child}. 
Catastrophic
forgetting and CL have received considerable attention
in computer vision
\cite[e.g.,][]{Zenke2017:ContinualLT,kirkpatrick2017:overcoming}, but 
far less attention within Natural Language Processing (NLP). 


\begin{figure}
\includegraphics[width=1.0\columnwidth]{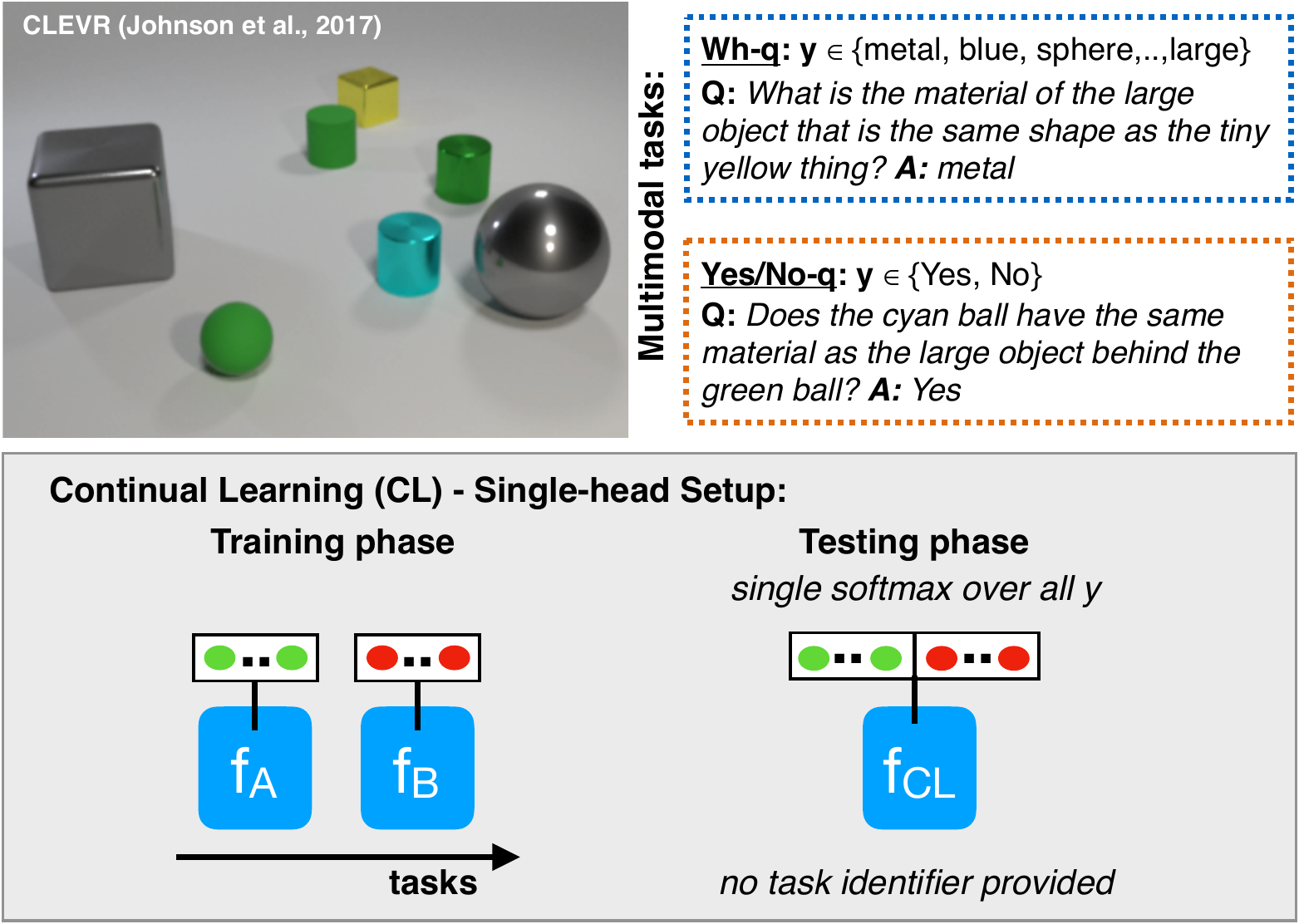}
\caption{Overview of our linguistically-informed CL setup for VQA.}	
\label{fig:overview}
\end{figure}

We investigate catastrophic forgetting in the context of multimodal models for Visual Question Answering~\cite{anto:vqa15} motivated by evidence from psycholinguistics. VQA is
the task of answering natural language questions about an image. 
Evidence from child language acquisition indicates that children 
learn Wh-questions before polar (Yes/No) questions~\cite{MoradlouGinzbug2016,MoradlouEtal2018}. 
Motivated by this finding, we design a set of linguistically-informed experiments: 
i) to investigate whether the order in which children acquire question types 
facilitates continual learning for computational models and, accordingly, the impact of \textit{task order} on catastrophic forgetting; 
ii) to measure how far two well-known CL approaches help to overcome the problem~\cite{robins1995:catastrophic,kirkpatrick2017:overcoming}\footnote{Code and data are available at the link \url{http://continual-vista.github.io/}.}. 


%
%

\paragraph{Contributions:} Our study contributes to the literature on CL in NLP. In particular:
i) we introduce a CL setup based on
linguistically-informed task pairs which differ with respect to question types and level of difficulty; 
ii) we show the importance of task order, an often overlooked aspect, and observe asymmetric synergetic effects;
iii) our results show that our VQA model suffers from extreme forgetting; rehearsal gives better results than a regularization-based method. Our error analysis shows that the latter
approach encounters problems even in discerning Task A after having
been trained on Task B. Our study opens the door to deeper
investigations of CL on linguistic skills with
different levels of difficulty based of psycholinguistics findings.

%
%
%
%



\section{Task Setup}\label{sec:tasks}
As a first step towards understanding the connection between 
linguistic skills and the impact on CL, we design  a set of experiments within VQA where tasks differ with respect to the \textit{type of question} and the \textit{level of difficulty} according to the psycholinguistics literature. The overall setup is illustrated in Figure~\ref{fig:overview} and  described next.
\paragraph{Dataset} 
CLEVR~\cite{johnson2017:clevr} allows to study the ability of VQA agents. It requires compositional language and basic spatial reasoning skills. 
Every question in CLEVR is derived by a Functional Program (FP) from a scene graph of the associated image. 
The scene graph defines the objects and attributes in the image. The FP contains functions corresponding to skills, 
 e.g., querying object attributes or comparing values (see 
Fig.~\ref{fig:overview}, upper).
Questions are categorized by their type. 
CLEVR consists of five question
  types  whose answer labels range over 15 attributes, 10 numbers, and
  ``yes''/``no'' (in total 27 labels).



\paragraph{Multimodal Tasks} 

We select the CLEVR sub-tasks `query\_attribute' and `equal\_attribute' with attributes {\em color, shape, material}, and {\em size}.
The two types of questions differ by answer type $y \in \mathcal{Y}$: 

\begin{itemize}[leftmargin=11pt,itemsep=0pt,topsep=0pt]
\item \textbf{Wh-questions} (Wh-q): Questions  about the  \textit{attribute} of an object, e.g., ``What is the material of the large object\ldots?'', where $y \in \{blue, cube, small, \ldots,
metal\}$ spans over $|color|=8$, $|shape|=3$, $|size|=2$ and $|material|=2$ (in total $|\mathcal{Y}|=15$).

\item \textbf{Yes/No questions} (Y/N-q): 
Questions that \textit{compare} objects with respect to an
attribute, e.g., ``Does the cyan ball have the same
material as \ldots?'', with $y \in \{ yes, no \}$ (in total $|\mathcal{Y}|=2$).
\end{itemize}

\paragraph{Task Order} 
We learn Task A followed by Task B ({\sc TaskA$\rightarrow$TaskB}), but experiment with \textit{both} directions, i.e., 
by first assigning Wh-q to Task A and Y/N-q to Task B, and vice versa. 
We expect that the inherent difficulty of a task and the order in which tasks are learned have an impact on CL.


\paragraph{Single-head Evaluation} CL methods can be tested in two ways. We opt for a {\em single-head} evaluation setup (see Fig.~\ref{fig:overview}, lower) with an output space over labels for all tasks (here: all CLEVR labels). In contrast, in a \textit{multi-head} setup predictions are restricted to task labels, as the task identifier is provided. Single-head is more difficult yet more realistic~\cite{chaudhry2018:riemannian}.  

\section{Models and Experiments} 
\label{sec:experiments}

\paragraph{VQA Model}  We take the model proposed
by~\newcite{yang2016:stacked} as a starting point, using  the code released
by~\newcite{john:infe17} (LSTM+CNN+SA).
Questions are encoded with a recurrent
neural network with Long Short-Term Memory (LSTM) units. Images are encoded with a
ResNet-101 Convolutional Neural Network (CNN) pre-trained on ImageNet \cite{he2016:deep}.
The two representations are combined using Spatial
Attention (SA)~\cite{yang2016:stacked} to focus on the most salient objects and properties in the image and text. The final answer distribution 
is predicted with a Multilayer Perceptron (MLP).  

\paragraph{Baselines}  


In order to measure catastrophic forgetting, 
we first consider per-task baselines: A random baseline (i.e., random stratified sample of the label distribution per task)
and the results of a model trained independently on each task 
(i.e., over task-specific $\mathcal{Y}$).
For CL, we report again a random baseline (this time a random stratified sample drawing predictions according to the answer distribution of both tasks), and 
we consider the \textit{Naive} and \textit{Cumulative} baselines proposed by \newcite{maltoni2018:continuous}.
The \textit{Naive} model is fine-tuned across tasks: It is first trained on Task
A and then on Task B starting from the
previously learned parameters. 
The \textit{Cumulative} model is trained from scratch on the training sets of both Task A and Task B.
This is a kind of upper bound, or performance that a CL model should achieve. 

\paragraph{Continual Learning Models} 
In CL there are two broad families of methods: Those that assume memory and access to explicit previous knowledge (instances), and those that have only access to compressed knowledge, such as previously learned parameters. These two families correspond to 
rehearsal and regularization, respectively. 
A widely-used regularization-based approach is \textit{Elastic Weight
Consolidation}~\cite[\emph{EWC},][]{kirkpatrick2017:overcoming}. A
regularization term, parametrized by $\lambda$, is added to the loss function aiming the model
to converge to parameters where it has a low error for both
tasks.
In the \emph{Rehearsal} approach~\cite{robins1995:catastrophic}, the
model is first trained on Task A, then the parameters are fine-tuned through
batches taken from a dataset containing a small number of examples of Task A and the training set of Task B. 
The selection of training examples of Task A is done
through uniform sampling.

\paragraph{Data and Training Details} 

Since CLEVR has no published ground-truth answers for the test set, we split the original validation set into a
validation and a test set.  To avoid performance impact due to different training data sizes, we 
downsample the training sets to the same size (Y/N-q data size), resulting in 125,654 training instances per task.
The validation and test sets contain, respectively, 26,960 and 26,774 data points for Wh-q and 13,417 and 13,681 data points  for Y/N-q. 

For the baselines, we select the model which reaches maximum
accuracy on the validation set of each task. For CL, we choose the
model with the highest CL score computed according to the validation
set of each task pair.
Details on hyper-parameters and evaluation metrics are provided in the supplementary material (SM).

\section{Results and Analysis}\label{ref:results}

The main results are provided in Table~\ref{tab:results}. There are several take-aways.

\paragraph{Task Difficulty} 
The results of the per-task models (cf.\
first two rows in Table~\ref{tab:results}) show that there is 
a large performance gap between the two tasks.
Wh-q is easier (.81) than Y/N-q (.52), regardless of the
fact that a priori the latter should be easier (as shown by the respective task-specific random baselines). 
The Y/N-q task-specific model  performs only slightly above chance (.52, in line with what 
\newcite{johnson2017:clevr} report for `equal\_attribute' questions).
This shows that despite 
the limited output space of the Y/N-q task, such type of questions in CLEVR
are complex and require reasoning skills~\cite{johnson2017:clevr}.

\paragraph{Catastrophic Forgetting} 
We observe that extreme forgetting is at play.
\emph{Naive} forgets the previously learned skill
completely: When tested on Task A after having been fine-tuned on Task
B, it achieves 0.0 accuracy on the first task \textit{for both
directions} (\textsc{I} and \textsc{II}, cf.\ Table~\ref{tab:results} lower). 
The \emph{Cumulative} model by nature cannot forget, since it is trained
on both tasks simultaneously, achieving 
.81 and .74 on Wh-q and Y/N-q, respectively. Interestingly, 
we observe an \textit{asymmetric synergetic effect}. 
Being exposed to the Wh-q task helps the \emph{Cumulative} model improve on Y/N-q, 
reaching results beyond the task-specific model (from .52 to .74).
The effect is not symmetric as the accuracy on Wh-q does not further increase.

\begin{table}
\resizebox{\columnwidth}{!}{
 \begin{tabular}{lcr|cr}\toprule
Random (per-task)  &  \multicolumn{2}{c}{\textsc{Wh:  0.09} } &  \multicolumn{2}{|c}{\textsc{Y/N: 0.50} } \\ 
LSTM+CNN+SA  &  \multicolumn{2}{c}{\textsc{Wh:}  0.81} &  \multicolumn{2}{|c}{\textsc{Y/N 0.52} }\\ 
 \midrule \toprule
\textsc{CL setups:}   &  \multicolumn{2}{c}{{\sc I)
                                             Wh$\rightarrow$Y/N}} &
                                                                   \multicolumn{2}{|c}{{\sc
                                                                         II)
                                                                         Y/N$\rightarrow$Wh}}\\
 &   Wh & Y/N &  Y/N &   Wh\\\hline
Random (both tasks)   & 0.04 & 0.25 &  0.25 & 0.04\\  
Naive &0.00  & 0.61 &  0.00
                                                            & 0.81\\
\hline
EWC   & 0.25  & 0.51  & 0.00 &
                                                                   0.83\\
Rehearsal & 0.75& 0.51 & 0.51
                                                                 &
                                                                          0.80 \\\hline
Cumulative   & 0.81 & 0.74  & 0.74  & 0.81 \\ 

\end{tabular}
}
\caption{Mean accuracy over 3 runs: Trained on each task independently
  (first two rows; per-task label space $\mathcal{Y}$) vs.\   CL setups (single-head label space over all $\mathcal{Y}$).}\label{tab:results}
\end{table}

\paragraph{Does CL Help?}
Current CL methods show only limiting (or no) effect.
\emph{EWC} performs bad overall: 
In the \textsc{II)} setup ({\sc y/n$\rightarrow$wh}, harder task first), \emph{EWC} does not yield any improvement over the {\em Naive} model; 
in the {\sc wh$\rightarrow$y/n} setup, the model's result on Task A is above chance level (.25 vs.~.04) but far off per-task performance (.81). 
The {\em Rehearsal} model forgets less than \emph{Naive} and \emph{EWC} in both setups:
In the {\sc y/n$\rightarrow$wh} setup, 
it is above chance level (.51 vs.\ .25) reaching per-task random baseline results on \textsc{Y/N} questions (i.e., the
model is able to identify Task A, despite the harder single-head setting, in contrast to the \emph{Naive} and \emph{EWC}
models). 
There is no boost derived from being exposed to the Wh-q 
task in any of the two setups. 

\paragraph{Task Order} The results in Table~\ref{tab:results} show that 
\textit{the order of tasks plays an important role}: {\sc wh$\rightarrow$y/n} facilitates CL
more than the opposite order: less forgetting is at place when {\sc wh} is learned first. 
This  confirms psycholinguistic evidence. Overall,
{\em Rehearsal} works better than \emph{EWC}, 
but mitigates forgetting only to a limiting degree.

\begin{figure*}
	\centering
	\includegraphics[width=1.0\linewidth]{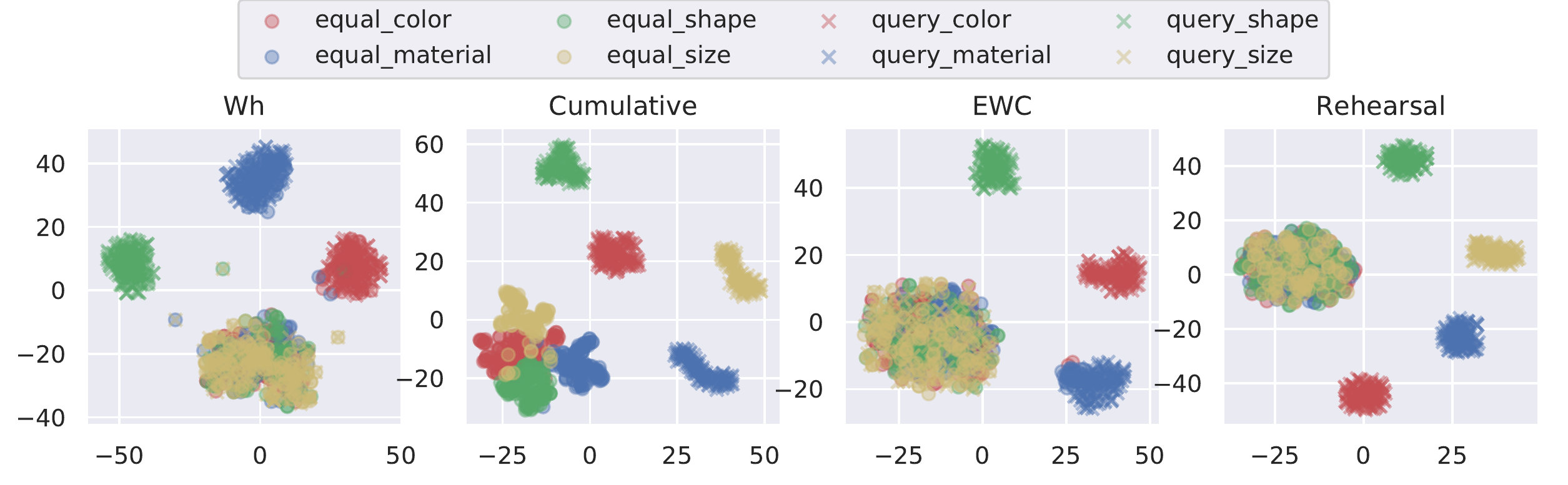}
	\caption{Analysis of the neuron activations on the penultimate
          hidden layer for the {I) \sc Wh $\rightarrow$ Y/N} setup. ``equal\_\{shape,color,material,size\}'' refers to \textsc{Y/N}-q, ``query\_\{..\}'' refers to \textsc{Wh}-questions.}
	\label{fig:clusters}
\end{figure*}

\paragraph{Analysis}
To get a deeper understanding of the models, we analyze the penultimate hidden layer on a sample of 512 questions from the test sets of both tasks (cf.\ Fig.~\ref{fig:clusters}) and relate the representations to confusion matrices of the whole test sets (provided in the SM) and test results (Table~\ref{tab:results}). 

First of all, the model trained on Wh-q discriminates Wh-questions about different attributes very well, reflected in overall high accuracy (.81). It
otherwise clusters all instances from the other task (Y/N-q, which it has not been trained on) around Wh-questions related to size. 

The \emph{Cumulative} model, in contrast, is able to further tease the different kinds of Y/N questions apart. Questions about different attributes become distinguishable in the plot, although overall Y/N questions remain  closer together than the clusters for Wh-q. This is in line with the lower
performance of \emph{Cumulative} on Y/N-q. 
Our examination of the  confusion matrices 
confirms that the two question types are never confused by the \emph{Cumulative} model. 
{In contrast, the \emph{Naive} model is very prone to this type of mistake (see plot in SM).
 
As for the CL models, Fig.~\ref{fig:clusters} (two rightmost plots) shows that \emph{EWC} 
learns representations which are rather similar to those learned by the model trained
on Wh-q independently: 
Y/N questions result in a big hard-to-distinguish ``blob'', and are confused with 
Wh-q about size, as visible in Fig.~\ref{fig:clusters} and the confusion matrix analysis (in the SM).
In contrast,  \emph{Rehearsal}  remembers how to distinguish
among all kinds of Wh-q \textit{and} between Wh-q and Y/N-q. 
The error analysis confirms that the model hardly makes any mistakes 
related to task confusion. However, despite the higher performance than \emph{EWC}, 
 \emph{Rehearsal} is still not able to discern well between different kinds of Y/N-q.
%




\section{Related Work}
\label{sec:related-work}

Early work on life-long learning~\cite{chen:life,mitchell2015} is related to ours, but typically concerns a single task (e.g., relation extraction). 
\newcite{lee:towa17} aims to transfer conversational skills from a synthetic domain to a customer-specific application in dialogue agents, while 
\newcite{yogatama2019:learning}  show that current models 
for different NLP tasks are not able to properly reuse previously learned knowledge. 


In general, continual learning has been mostly studied in computer vision. 
To the best of our knowledge, little has been done on catastrophic
forgetting in VQA.
A study on forgetting in the context of VQA and closest to ours is~\newcite{perez2018:film}. They show  
that their model forgets after being fine-tuned
on data including images with objects of colors other than those
previously seen. We took this work as starting point and extended it to
consider different types of questions 
and to test different CL methods beyond fine-tuning.


\section{Conclusion}
\label{sec:conclusion}

We assessed to what extent a multimodal model suffers from
catastrophic forgetting in a VQA task. We built two tasks involving
different linguistic characteristics which are known to be learned
sequentially by children and on which multimodal models reach
different performance. 

Our results show that dramatic forgetting is at
play in VQA, and for the tested task pairs we empirically found {\em
 Rehearsal} to work better than a regularization-based method ({\em EWC}). 
 More importantly, we show that the \textit{order} in which models learn
tasks is important, {\sc wh$\rightarrow$y/n} facilitates continual learning
more than the opposite order, thereby confirming psycholinguistic evidence. 
 
 Our 
 error analysis 
highlights the importance of taking the kind of mistakes made by the models  into account:
 A model that does not detect Task A after having been
exposed to Task B should be penalized more than a model that answers
Task A with wrong task-related labels, but is still capable of identifying the task.
Most importantly, our study revealed that differences in the inherent  
difficulty of the tasks at hand 
can have a strong impact on continual learning. Regularization-based
methods like {\em EWC} appear to work less well when applied to tasks
with different levels of difficulty, as in our experiments. 
We reserve a deeper investigation of this aspect to future research.

\section*{Acknowledgements}
We kindly acknowledge the support of NVIDIA
 Corporation with the donation of the GPUs used in our research to the University of Trento and IT University of Copenhagen. 
R.~Fern\'{a}ndez was funded by the Netherlands Organisation for Scientific Research (NWO) under VIDI grant nr.~276-89-008, {\em Asymmetry in Conversation}.  

\bibliography{../bibliography}
\bibliographystyle{acl_natbib}




\end{document}